\pdfoutput=1
\documentclass[conference, a4paper]{IEEEtran}
\IEEEoverridecommandlockouts
\usepackage{cite}
\usepackage{amsmath,amssymb,amsfonts}
\usepackage{algorithmic}
\usepackage{graphicx}
\usepackage[font=footnotesize, justification=centering]{caption}
\usepackage{textcomp}
\usepackage{xcolor}
\usepackage{url}
\usepackage{tikz}
\usepackage{titlesec}
\titlespacing*{\section}{0pt}{8pt}{4pt}
\titlespacing*{\subsection}{0pt}{6pt}{3pt}
\titlespacing*{\subsubsection}{0pt}{4pt}{2pt}
\usetikzlibrary{arrows.meta, shapes, automata, positioning, fit, backgrounds}
\usepackage{geometry}
\def\BibTeX{{\rm B\kern-.05em{\sc i\kern-.025em b}\kern-.08em
    T\kern-.1667em\lower.7ex\hbox{E}\kern-.125emX}}
\begin{document}

\title{Spatially-Aware Self-Organizing Graph Autoencoders via Localized Hebbian Learning}

\makeatletter
\def\@IEEEauthorblockNfont{\fontsize{11pt}{13pt}\selectfont}
\def\@IEEEauthorblockAfont{\fontsize{10pt}{12pt}\selectfont}
\makeatother

\author{
\IEEEauthorblockN{Robin Koç\thanks{Our PyTorch implementation and interactive evaluation dashboard are publicly available at \protect\url{https://github.com/robinkoc/LooMs}.}}
\IEEEauthorblockA{\textit{Computer Engineering Department} \\
\textit{Middle East Technical University}\\
Ankara, Türkiye \\
robin.koc@metu.edu.tr}
\and
\IEEEauthorblockN{Hazal Moğultay Özcan}
\IEEEauthorblockA{\textit{Computer Engineering Department} \\
\textit{Middle East Technical University}\\
Ankara, Türkiye \\
mogultay@metu.edu.tr}
\and
\IEEEauthorblockN{Fatoş T. Yarman Vural}
\IEEEauthorblockA{\textit{Computer Engineering Department} \\
\textit{Middle East Technical University}\\
Ankara, Türkiye \\
yarman@metu.edu.tr}
}

\maketitle

\begin{abstract}
Traditional autoencoders rely on static network architectures to learn latent representations. Recently we suggested a dynamic architecture inspired by Hebbian plasticity, enabling topological self-organization among convolutional feature maps \cite{koc_loom}. Early implementations of this paradigm infer relationships among feature maps using global average pooling and model their interactions through scalar-valued connections. This approach  falls short to fully exploit the underlying spatial information, embedded in input images. Furthermore, abrupt transitions between different loss functions and training dynamics can introduce representation shocks, leading to unstable feature learning. In this paper, we introduce a Spatially-Aware Self-Organizing Graph Autoencoder (S-LooM) to address these structural bottlenecks. By using spatial cross-correlation, and spatial kernels, S-LooM establishes dynamic connections among feature maps. We replace static scalar message passing of autoencoders with learnable spatial kernels and eliminate high-frequency deconvolutional artifacts observed in decoders using nearest-neighbor upsampling. Via a soft-blending training curriculum, the latent topology aligns smoothly during training. Our experiments show that S-LooM yields a more stable topology, compared to the state of the art dynamic architectures. Meaning, the graph structure that emerge has more resistance to feature entanglement and noise. It forms  functional communities responsible for  visual attributes, such as illumination, shape  and  texture even under high connection densities.
\end{abstract}

\renewcommand{\IEEEkeywordsname}{Keywords}
\begin{IEEEkeywords}
\textit{graph neural networks, feature extraction, Hebbian plasticity, self-organization, spatial awareness}
\end{IEEEkeywords}

\section{Introduction}
Deep learning architectures, such as autoencoders, capture the underlying structure of given datasets through convolutional encoders and decoders. Although they are inspired by biological neurons from the start, they usually rely on static networks designed prior to training. This contrasts with biological neural networks, which exhibit adaptive connectivity through Hebbian plasticity, where synapses evolve dynamically based on co-activation \cite{hebb1949} and exhibit highly adaptive plasticity.

In a  foundational study \cite{koc_loom}, we introduced the LooM architecture, bridging the gap between the biological principle and autoencoders, by dynamically optimizing latent connectivity using a self-organizing graph. While LooM provided a proof-of-concept for Hebbian-guided representation learning, its initial implementation revealed specific bottlenecks in translating this concept to convolutional features:
\begin{enumerate}
    \item \textbf{Global Averaging in Hebbian Statistics Calculation:} Determining Hebbian co-activation by averaging entire feature maps can merge features that exist in completely disjoint regions of the visual input, and provide incorrect co-activation statistics.
    \item \textbf{Restrictive Scalar Message Passing:} Relying only on single scalar weights to pass messages between latent nodes limits structural expressivity, because a scalar can only uniformly amplify or dim an entire feature map. This was pushing the model to silencing or activating entire feature maps rather than allowing them to cooperate with one another.
    \item \textbf{Curriculum Transitions:} Even though the introduction of graph loss value is gradual, abrupt introduction of message passing can cause shocks, forcing the decoder to unlearn weights as latent distributions suddenly shift.
    \item \textbf{Generative Checkerboard Artifacts:} Standard transposed convolutions often suffer from uneven kernel overlaps, which introduces high-frequency checkerboard artifacts to the spatial outputs. Since the Hebbian teacher relies on statistical correlation, this artifacts act as noise when gathering Hebbian statistics, forcing the model into forming false topological connections.
\end{enumerate}

To address these challenges, we propose \textbf{S-LooM} (Spatially-Aware LooM), a graph-autoencoder architecture driven by Hebbian learning. The S-LooM architecture is built upon three main components: a convolutional encoder, a convolutional decoder, and a self-organizing graph layer between them where each node represents a feature map extracted by the encoder. This graph structure evolves during training to discover relationships between feature maps, and allow them to interact with each other during later parts of the training and inference. The graph layer relies on two core mechanisms, Hebbian statistics gathering by using co-activations between feature maps, and message passing for cooperation between said feature maps based on gathered statistics. To overcome the limitations of early implementations, S-LooM replaces the scalar message passing between nodes with learnable spatial kernels that allow spatial feature blending, and utilizes localized spatial cross-correlation to ensure that connections only form between features co-activating within similar receptive fields. Finally, to stabilize this dynamic learning process, we employ an artifact-free decoding scheme to prevent artificial noise from corrupting the Hebbian statistics, alongside a soft-blending curriculum to gradually introduce message passing and mitigate representation shock. To promote reproducibility, our PyTorch implementation and the interactive evaluation dashboard are publicly available at \url{https://github.com/robinkoc/LooMs}.

\section{Related Work}

\noindent \textbf{Dynamic Latent Topologies:} 
Previous studies on representation learning and Graph Neural Networks (GNNs) generally assume static, predefined adjacency matrices \cite{kipf2016gae}. However recent studies has focused on dynamic graph structure learning \cite{zhu2021}. Finding efficient yet adaptive mechanisms to discover latent feature relationships on the fly remains a challenge.

\noindent \textbf{Hebbian Plasticity in Modern Architectures}
Hebbian learning served as the foundation for associative memory in Hopfield networks \cite{hopfield1982},  but its \textbf{non-differentiable} nature has limited its integration into backpropagation-driven deep learning \cite{miconi2018}. In a recent study \cite{koc_loom}, we bridged this gap by using Hebbian co-activation statistics as a detached topological teacher for a learnable autoencoder graph. However, proposed model relied on global average pooling to gather co-activation statistics, abstracted away the spatial geometric context, and treated activations of complex two-dimensional feature maps as isolated scalars.

\noindent \textbf{Spatial Coherence and Generative Artifacts}
A persistent goal in visual generative modeling is the preservation of visual integrity. Many traditional autoencoder architectures flatten latent variables into dense vectors, destroying spatial dimensions. And models that retain the spatial nature of feature maps, utilize transposed convolutions, thus frequently suffer from checkerboard artifacts caused by kernel overlaps. As demonstrated by Odena et al. \cite{odena2016}, replacing standard deconvolution operations with resize-convolutions (nearest-neighbor upsampling followed by standard convolution) eliminates these high-frequency distortions. S-LooM requires the integration of this artifact-free decoding since artificial structural noise corrupts the Hebbian co-activation statistics that guide the model itself.

\section{General Architecture of S-LooM}

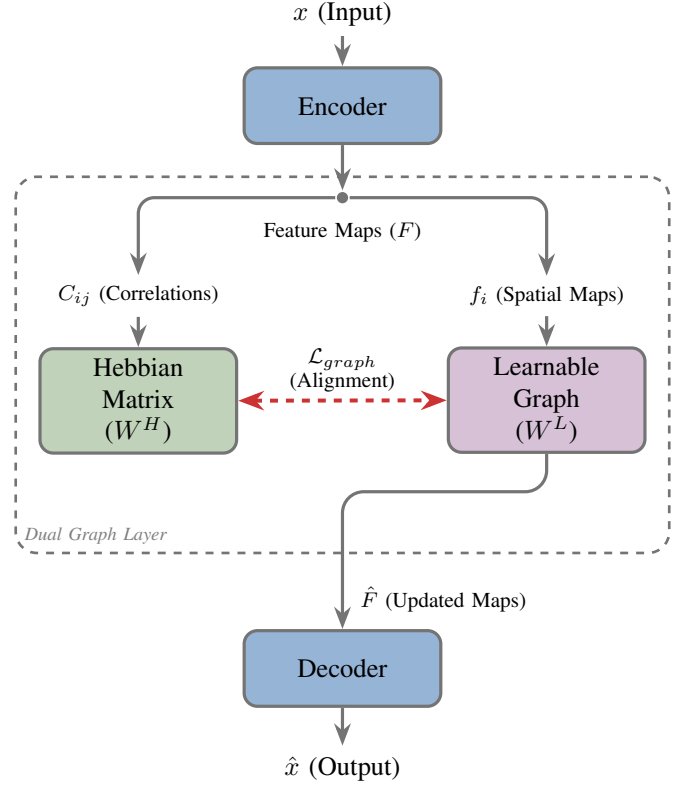
\begin{figure}[htbp]
    \centering
    \begin{tikzpicture}[
        >={Stealth[scale=1.0]}, 
        line width=1.2pt,       
        define rgb/.code={\definecolor{myblue}{RGB}{150,180,214}
                          \definecolor{mygreen}{RGB}{192,211,186}
                          \definecolor{mypurple}{RGB}{215,193,215}
                          \definecolor{mygray}{RGB}{116,116,116}
                          \definecolor{myred}{RGB}{205,50,50}},
        define rgb,
        draw=mygray,
        box/.style={rectangle, rounded corners=5pt, minimum width=2.6cm, minimum height=1.0cm, align=center, draw=mygray, font=\normalsize, text=black},
        bluebox/.style={box, fill=myblue},
        greenbox/.style={box, fill=mygreen},
        purplebox/.style={box, fill=mypurple},
        txt/.style={text=black, font=\normalsize},
        smalltxt/.style={text=black, font=\footnotesize, align=center},
        dot/.style={circle, fill=mygray, inner sep=1.5pt, draw=none}
    ]


    \node[txt] (input) {$x$ (Input)};
    
    \node[bluebox, below=0.4cm of input] (enc) {Encoder};
    
    \node[dot, below=0.6cm of enc] (split) {};
    \node[txt, below=0.1cm of split, font=\footnotesize] {Feature Maps ($F$)};
    
    \node[smalltxt] (act) at ([xshift=-2.7cm, yshift=-1.3cm]split) {$C_{ij}$ (Correlations)};
    \node[smalltxt] (maps) at ([xshift=2.7cm, yshift=-1.3cm]split) {$f_i$ (Spatial Maps)};
    
    \node[greenbox, below=0.4cm of act] (hebb) {Hebbian\\Matrix\\($W^H$)};
    \node[purplebox, below=0.4cm of maps] (learn) {Learnable\\Graph\\($W^L$)};
    
    \path (hebb) -- (learn) node[midway, smalltxt, yshift=0.4cm] (loss) {$\mathcal{L}_{graph}$\\(Alignment)};
    \draw[<->, dashed, myred, line width=1.5pt] (hebb) -- (learn);
    
    \node[bluebox, below=6.4cm of enc] (dec) {Decoder};
    
    \node[txt, below=0.5cm of dec] (out) {$\hat{x}$ (Output)};
    
    \draw[->] (input) -- (enc);
    \draw[->] (enc) -- (split);
    
    \draw[->, rounded corners=6pt] (split) -| (act);
    \draw[->, rounded corners=6pt] (split) -| (maps);
    \draw[->] (act) -- (hebb);
    \draw[->] (maps) -- (learn);
    
    \draw[->, rounded corners=8pt] (learn.south) -- ++(0,-0.5) coordinate (j1) -- (j1 -| dec.north) -- (dec.north);
    
    \node[smalltxt, right=0.1cm] at ([yshift=0.4cm]dec.north) {$\hat{F}$ (Updated Maps)};
    
    \draw[->] (dec) -- (out);
    
    \node[draw=mygray, dashed, line width=1pt, rounded corners=8pt, inner xsep=0.3cm, inner ysep=1.3cm, fit=(act) (maps) (hebb) (learn) (loss)] (module) {};
    
    \node[above right, text=gray, font=\scriptsize\itshape] at (module.south west) {Dual Graph Layer};
    
    \end{tikzpicture}
    \caption{Graphical abstract of the S-LooM architecture. The encoder extracts feature maps, which branch into localized spatial correlations ($C_{ij}$) to be used for the Hebbian Matrix ($W^H$), and spatial maps themselves ($f_i$) are fed into the Learnable Graph ($W^L$). The red arrow between the graph structures represents the topological alignment loss.}
    \label{fig:graphical_abstract}
\end{figure}

Following the baseline established in the LooM framework \cite{koc_loom}, the S-LooM architecture is composed of three primary parts: a convolutional encoder that extracts spatial feature maps, and a convolutional decoder that reconstructs the original input from the feature maps and between them a dual-graph layer that orchestrates these features topologically, as illustrated in Fig. \ref{fig:graphical_abstract}. This modular pipeline keeps topological self-organization in the latent space, preserving the foundational encoding and decoding mechanics of the autoencoding process.

The S-LooM encoder $E$ produces $K$ feature maps that are explicitly preserved as two-dimensional matrices. This design choice retains the geometric context for each input image $x \in \mathbb{R}^{H \times W \times C}$. A \textit{sigmoid} function is applied at the output of the encoder to normalize the value ranges. The matrix $F = E(x)$ containing the extracted feature maps is defined as:

\begin{equation}
    F = \{f_1, f_2, \dots, f_K\}, \quad f_i \in [0,1]^{H' \times W'}. \label{eq: 1}
\end{equation}

These feature maps are then passed to a graph layer $G$ inserted between the encoder and the decoder. This layer dynamically reorganizes the feature maps based on their emerging co-activation relationships, strictly without altering their dimensions:

\begin{equation}
\hat{F} = G(F), \quad \hat{F} = \{\hat{f}_1, \hat{f}_2, \dots, \hat{f}_K\}, \quad \hat{f}_i \in \mathbb{R}^{H' \times W'}.
\end{equation}

The decoder $D$ then reconstructs the image $\hat{x}$ from these modified matrices:

\begin{equation}
\hat{x} = D(\hat{F}), \quad \hat{x} \in \mathbb{R}^{H \times W \times C}.
\end{equation}

However, a structural vulnerability in the prior implementations was their reliance on standard transposed convolution operations in the decoder. Transposed convolutions with stride greater than one, frequently suffer from uneven kernel overlaps, which introduces periodic checkerboard artifacts into the reconstructed outputs. Because our Hebbian graph relies heavily on spatial cross-correlation between feature maps to learn the latent topology, artificial noise can create false statistical correlations by back-propagating from the decoder, distorting gathered Hebbian statistics, thus the graph structure itself.

To overcome this problem, the S-LooM uses artifact-free resize-convolutions instead of transposed convolutions in decoder. Let $h \in \mathbb{R}^{C \times H' \times W'}$ be any feature map within the decoder. First, we expand its spatial dimensions by a scale factor $s$ using nearest-neighbor upsampling, defined as $h^u_{c, i, j} = h_{c, \lfloor i/s \rfloor, \lfloor j/s \rfloor}$. This duplicates existing values to stretch the feature map without creating overlapping artifacts. A standard 2D convolution with learnable weights $W$ is then applied to process the features into the next layer's output: $l = h^{u} \ast W$.

By separating the spatial expansion and learnable convolution step ($W$), we make sure that every pixel has uniform structural paths during backpropagation. This prevents the artificial checkerboard patterns often caused by transposed convolutions, keeping the gradients clean so the Hebbian graph learns from real visual features rather than structural noise.

\section{Spatially-Aware Dual-Graph Layer}

The core of the S-LooM framework is the dual-graph layer. This layer orchestrates the topological self-organization without requiring a predefined adjacency matrix. It consists of two parallel structures: the \textbf{learnable autoencoder graph} and the \textbf{Hebbian graph}. The learnable autoencoder graph, $G^L$, is defined by a node set ($V^L$) and an edge weight matrix ($W^L$):

\begin{equation}
G^L = (V^L, W^L), \quad V^L = \{f_{1}, f_{2}, \dots, f_{K}\}.
\end{equation}

where $f_i$ are the spatial feature maps obtained in Equation \eqref{eq: 1}.

The Hebbian graph, represented by its edge weight matrix ($W^H$), acts as a topological guide. It dictates the statistical relationships between feature maps  based on their co-activations. As can be seen in the dashed red alignment arrow in Fig. \ref{fig:graphical_abstract}, this topological guidance is enforced on the learnable graph $G^L$ through an addition to the standard autoencoder loss function. By pulling the weights of learnable autoencoder graph toward the Hebbian graph during backpropagation, this mechanism ensures that the emergent graph reflects visual relationships, forming a latent space without sacrificing reconstruction accuracy.

Because the network's learnable parameters and the statistical co-activations are continuously updated throughout the learning process, both graphs evolve simultaneously. To formalize this, let $t$ represent the current training step. At any given time step $t$, the dual-graph layer $G(t)$ is expressed as the union of the evolving weights over the shared nodes:

\begin{equation}
G(t) = \{W^H(t), G^L(t)\},
\end{equation}

 In order to enforce genuine inter-feature message passing and prevent the network from collapsing into an identity mapping, the graph layer explicitly masks self-connections, $W_{ii} = 0$. Specifically, the connection that any feature node $i$ makes with itself is strictly zeroed out, enforcing $W_{ii}(t) = 0$ across both graphs at all time steps $t$.

\subsection{Localized Hebbian Co-activation}


In S-LooM, co-activation statistics are calculated according to localized spatial alignment. Given centered and normalized feature maps, the co-activation between features $f_i$ and $f_j$ at time $t$ is calculated directly via their maximum spatial cross-correlation also at time $t$:

\begin{equation}
C_{ij}(t) = \max_{u,v} (f_i(t) \star f_j(t))(u,v),
\end{equation}

where $\star$ denotes the cross-correlation operator, and $(u,v)$ represent the two-dimensional spatial shifts applied during the operation. 

Since S-LooM derives Hebbian updates directly from pairwise spatial similarities rather than independent node scalars, evaluating activation thresholds on individual nodes is mathematically incompatible. Instead, to explicitly enforce graph sparsity, S-LooM dynamically prunes weak connections by computing a quantile threshold directly over the \textbf{edges}. Based on a target sparsity ratio $p$, this approach zeroes out weak structural connections while accumulating the correlation values of the surviving active edges according to the following formula:

\begin{equation}
W^H_{ij}(t) = (1 - \alpha) W^H_{ij}(t-1) + \alpha \cdot C_{ij}(t) \cdot \mathbb{I}(C_{ij}(t) > \tau_t),
\end{equation}

where $\mathbb{I}$ is the indicator function that outputs 1 if the condition is met and 0 otherwise, and $\alpha \in (0,1]$ is a momentum coefficient controlling the update speed of the connections. Before used as a target for the learnable graph, the Hebbian graph is normalized to create a stable and balanced "graph skeleton" ($\hat{W}^H$):

\begin{equation}
\hat{W}^H = \frac{W^H}{\max_{m,n} (W^H_{m,n}) + \epsilon},
\end{equation}

where $m$ and $n$ iterate over all feature nodes to find the maximum edge weight across the entire Hebbian graph.

\subsection{Spatial Message Passing}

Message passing enables latent feature maps to exchange information and cooperate based on their co-activation relationships. However, relying on scalar weights to formulate these messages limits structural expressivity. When a message is computed by multiplying a source feature map with a single scalar, the operation uniformly amplifies or dims the entire map. To overcome this restriction, S-LooM introduces learnable spatial kernels between nodes. Let $k_{ji} \in \mathbb{R}^{s \times s}$ represent a learnable 2D convolutional kernel defining the spatial transformation of messages from feature node $j$ to node $i$. Unlike a scalar weight, a spatial kernel allows the model to shift features, allowing the network to construct cooperative visual representations by recombining parts of co-activating feature maps. To ensure the network initially relies entirely on the encoder feature map outputs without untrained graph interference, these spatial kernels are initialized from a zero-mean normal distribution with a small standard deviation (0.02).

During the forward pass, the continuous scalar edge weights $w_{ji} = \sigma(W^L_{ji})$ act as continuous gating mechanisms that determine the scales of the values in these spatial kernels. The total aggregated message $M_i$ passed to the $i$-th feature map is computed using a 2D convolution operator (denoted by $\ast$) between spatial kernel and source (j-th) feature map itself:

\begin{equation}
M_i = \sum_{j \neq i} (w_{ji} \cdot k_{ji}) \ast f_j
\end{equation}

To maintain numerical stability and prevent exploding activations during graph aggregation, S-LooM employs a strict scale-preservation mechanism. The updated feature map $\hat{f}_i$ is computed by adding the aggregated messages to the original feature map $f_i$ (acting as an identity skip connection) and normalizing by the total weight of the incoming edges:

\begin{equation}
\hat{f}_i = \frac{f_i + M_i}{1 + \sum_{j \neq i} w_{ji}}
\end{equation}

Because the initial encoder features $f_i$ are constrained to a $[0, 1]$ range via a sigmoid activation, this normalization formulation mathematically bounds the aggregated output. It ensures that even under highly dense graph topologies, the magnitude of the latent representations fed into the decoder remains stable and protected from exploding gradients.

\section{Soft-Blending Curriculum and Optimization}

The network jointly optimizes the visual reconstruction performance ($L_{recon}$) and the graph topology similarity ($L_{graph}$):

\begin{equation}
L_{recon}= \|x-\hat{x}\|_1, \quad L_{graph} = \frac{1}{K^2} \sum_{i,j} |\sigma(W^L_{ji}) - \hat{W}^H_{ji}|.
\end{equation}

The total loss is dynamically balanced as:

\begin{equation}
    L_{total} = L_{recon} + \gamma(t) \cdot L_{graph}.
\end{equation}

The LooM training protocol utilized a ramp function to smooth the transition between training phases, but this ramp function was only effective on the graph loss coefficient ($\gamma$). So, LooM still subjected the forward pass to a hard, half-instantaneous transition by allowing the message passing to take effect on the feature maps instantly (switching abruptly from $\hat{F} = F$ to $\hat{F} = G(F)$). This caused sudden shifts in the latent distribution fed to the decoder and caused representation shock. To resolve this, S-LooM refines the multi-stage curriculum by applying a soft-blending mechanism directly to the feature maps:

\begin{itemize}
    \item \textbf{Stage 1 (Warm-up):} The graph layer is bypassed ($\hat{F} = F$). The goal is merely for the convolutional filters to extract basic spatial features and reach a stable reconstruction error.
    \item \textbf{Stage 2 (Topological Alignment \& Soft-Blending):} The graph layer is activated. To prevent representation shock, S-LooM introduces a soft-blending curriculum function $\beta(t)$. Over a predefined ramp period of epochs, the latent representation smoothly interpolates between the raw encoder features $F$ and the graph-processed features $G(F)$ as follows:
    \begin{equation}
    \hat{F}_{blend} = (1 - \beta(t))F + \beta(t)G(F),
    \end{equation}
    where $\beta(t)$ linearly scales from $0$ to $1$. Simultaneously, the $\gamma$ coefficient scales up from zero to equalize gradient magnitudes, optimizing $W^L$ via $L_{graph}$. Basically not just gradually introducing the graph loss to the mix, also gradually increasing the effects of the message passing on the feature maps.
    \item \textbf{Stage 3 (Structural Specialization):} The graph layer is frozen. The model subjects the feature maps to a final fine-tuning on the fixed topology.
\end{itemize}

This smooth curriculum allows the decoder to adapt gradually to the changing latent topology, contributing to the stable validation performance observed in our empirical evaluations.

\section{Experiments and Results}

In this section, we evaluate the performance of S-LooM. We are constraining our evaluations to Cifar-10 dataset to be able to isolate the effects of architectural changes while benchmarking it against the original LooM architecture \cite{koc_loom}. Because this work builds upon the conceptual basis of a prior study, our evaluations are focused on how the shift from global scalar routing to spatially-aware message passing impacts baseline fidelity, stability across graph densities, and visual explainability.

\subsection{Experimental Setup and Baseline Performance}

To ensure a fair comparison, S-LooM was trained using identical hyper-parameters to the prior LooM study: a latent space of $K=32$ feature maps, the Adam optimizer (learning rate of 0.001), $L_1$ loss, and a 150-epoch limit. 

\begin{figure}[htbp]
    \centering
    \includegraphics[width=0.75\linewidth]{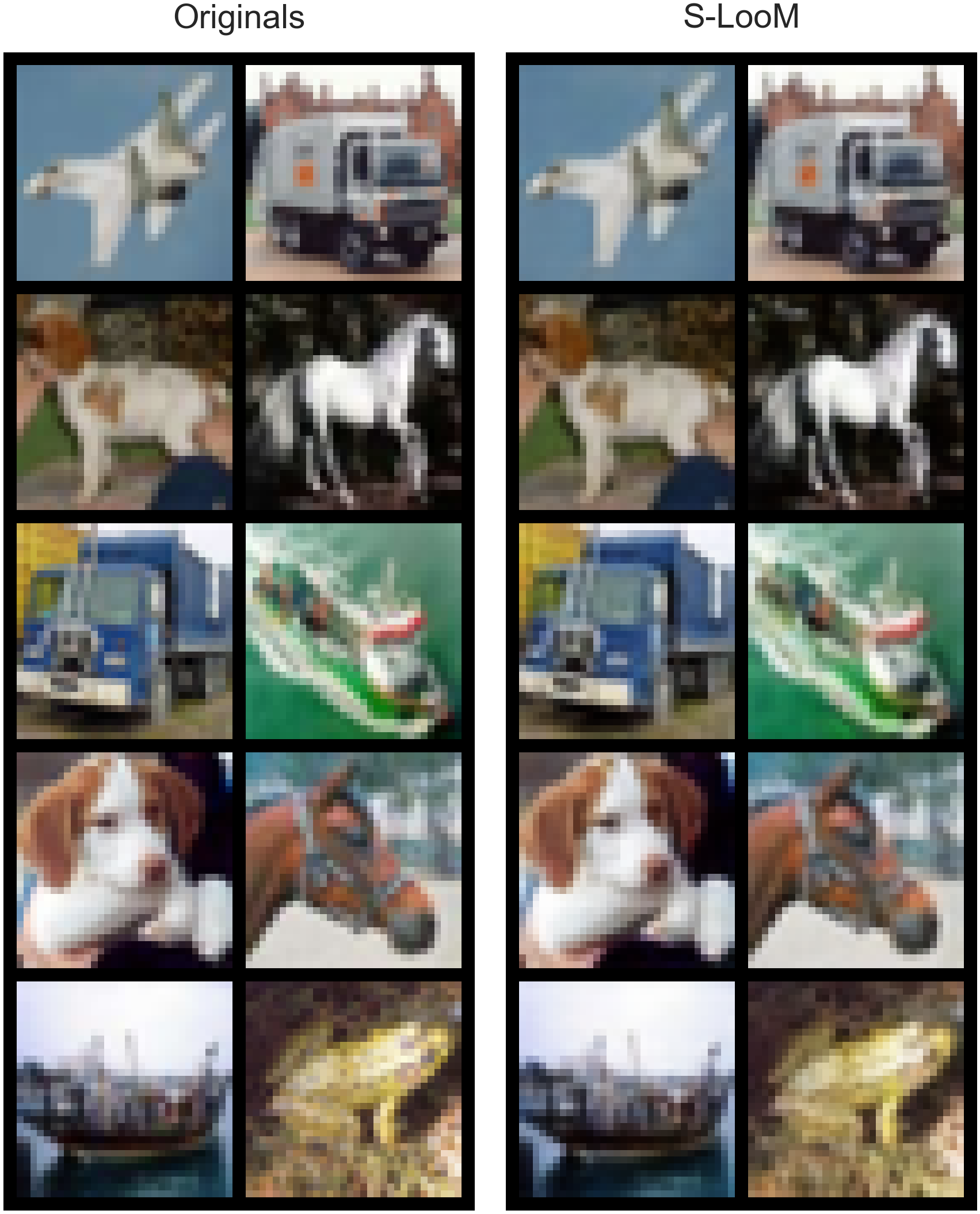}
    \caption{Qualitative evaluation of S-LooM reconstructions. The model yields high quality reconstructions while maintaining a strictly governed topological latent space.}
    \label{fig:reconstructions}
\end{figure}

Under a highly sparse target constraint (sparsity ratio of $p=0.05$), the prior LooM model recorded a validation loss of 0.0272. In contrast, S-LooM achieved a validation loss of \textbf{0.0114} under the exact same conditions. This reduction in reconstruction error validates the combination of architectural upgrades. The artifact-free nearest-neighbor upsampling provides the Hebbian teacher with clean spatial signals, while the integration of localized cross-correlation and learnable spatial kernels provides more expressivity than the prior architecture's rigid scalar message passing. As visually confirmed in Fig. \ref{fig:reconstructions}, this spatial approach successfully yields high-fidelity reconstructions while maintaining a strictly governed topological latent space.

\subsection{Stability Across Graph Densities and Feature Entanglement}

The target sparsity ratio $p$ determines the density of the topology dictated by the Hebbian teacher. Table \ref{tab:parameter_search} displays the validation losses obtained across a range of sparsity ratios.

\begin{table}[htbp]
\centering
\caption{COMPARISON OF TARGET SPARSITY ON VALIDATION LOSS}
\label{tab:parameter_search}
\begin{tabular}{|c|c|c|}
\hline
\textbf{Sparsity Ratio ($p$)} & \textbf{Prior LooM Loss} & \textbf{S-LooM Loss} \\ \hline
0.05 & 0.0272 & \textbf{0.0114} \\ \hline
0.10 & 0.0283 & \textbf{0.0103} \\ \hline
0.20 & 0.0305 & \textbf{0.0104} \\ \hline
0.30 & 0.0322 & \textbf{0.0107} \\ \hline
0.40 & 0.0340 & \textbf{0.0097} \\ \hline
0.50 & 0.0361 & \textbf{0.0106} \\ \hline
\end{tabular}
\end{table}

A notable limitation of the foundational architecture was its fragility to scaling; as the graph became denser ($p \ge 0.30$), the inclusion of statistically weaker connections introduced direct structural noise, steadily degrading reconstruction performance. S-LooM, however, maintains a highly stable validation loss ($\sim$0.010) across all densities.

\begin{figure}[htbp]
    \centering
    \includegraphics[width=1.0\linewidth]{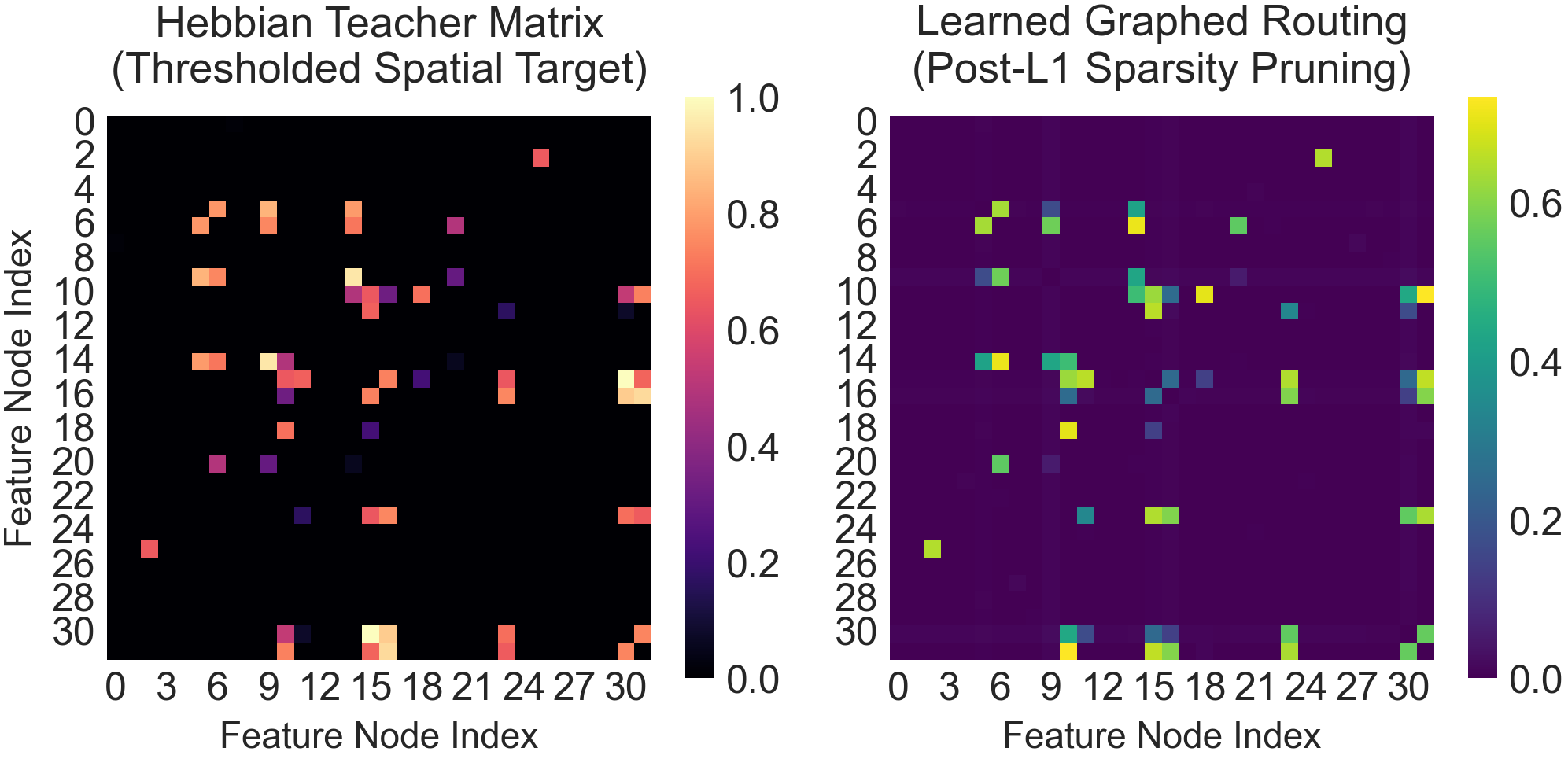}
    \caption{Matrix visualization comparing the converged learnable spatial graph and the localized Hebbian teacher graph with target  sparsity $p = 0.05$, demonstrating strong topological alignment.}
    \label{fig:matrix}
\end{figure}

However, qualitative analysis reveals that while the loss remains mathematically stable at high densities ($p=0.50$), the latent space suffers from severe feature entanglement. In dense configurations, the network is forced into a tangled state where ablating a single node causes widespread, image-wide color and shape distortions. Conversely, at high sparsity ($p=0.05$), the network exhibits functional modularity. As demonstrated in Fig. \ref{fig:matrix}, comparing the converged learnable spatial graph ($W^L$) to the localized Hebbian teacher ($W^H$) reveals a strong, highly sparse topological alignment. This alignment yields interpretable feature relationships without sacrificing the quantitative reconstruction performance.

\subsection{Functional Specialization and Community Emergence}

A primary goal of the foundational LooM framework was to organically cluster features into semantically meaningful communities.  S-LooM's spatial awareness comes closer to this vision. Applying community detection algorithms to the converged S-LooM spatial graph ($p=0.05$) reveals a highly specialized functional division. We continued with the Louvain method over alternatives because it organically discovers communities based on edge weights without requiring a predefined number of clusters, making it a natural fit for analyzing our self-organizing topology.

\begin{figure}[htbp]
    \centering
    \includegraphics[width=0.8\linewidth]{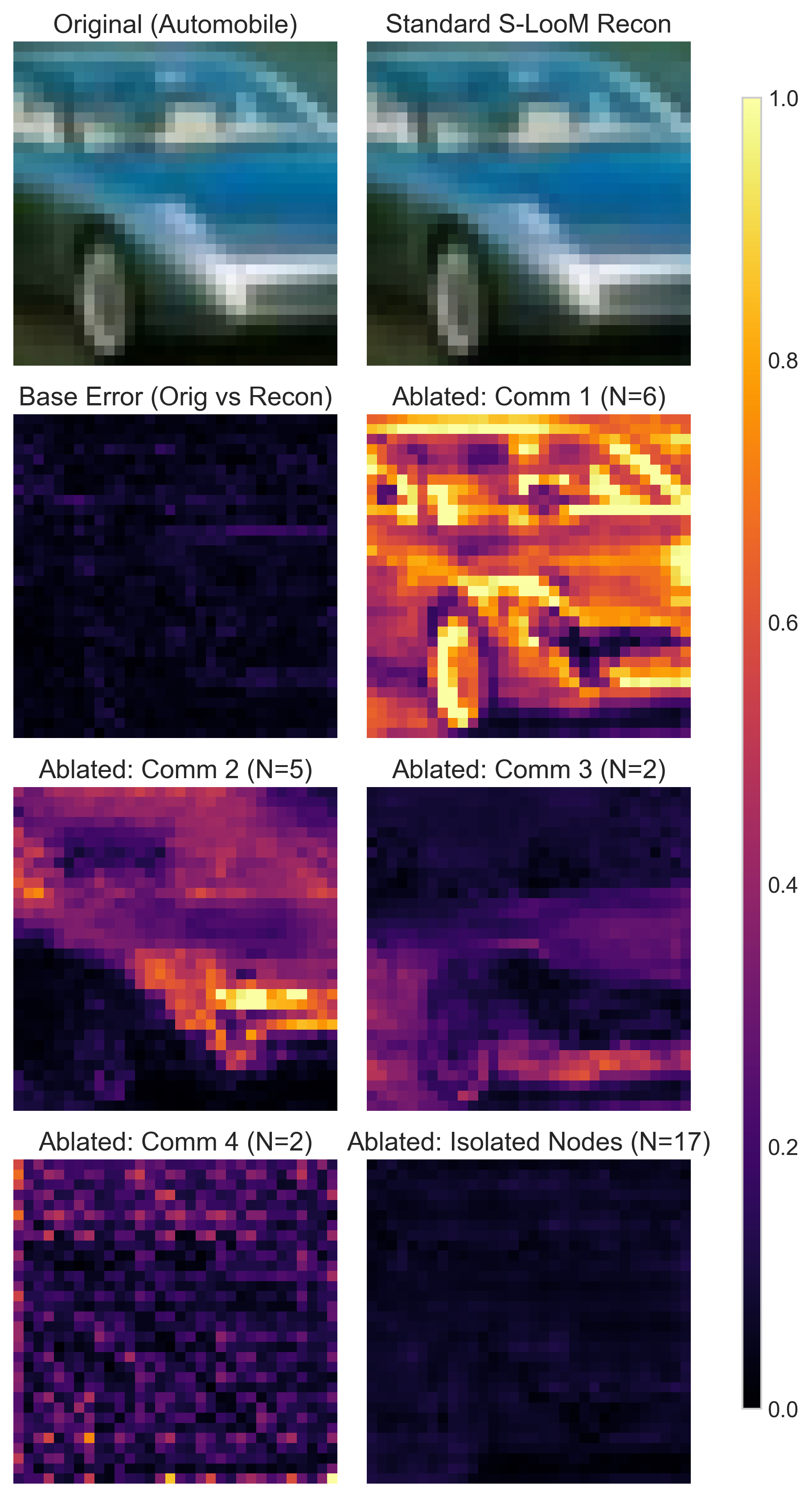}
    \caption{Community Ablation Analysis. The difference maps isolate the distinct semantic responsibilities of each Louvain-detected community. Note the minimal impact of the isolated nodes, demonstrating the graph's ability to quarantine uninformative filters.}
    \label{fig:communities}
\end{figure}

By systematically silencing all edges associated with specific clusters on the graph structure during inference, we can map the visual responsibility of each community. Fig. \ref{fig:communities} illustrates this ablation process on a selected validation sample. While this specific image provides a particularly clear visualization of the network's compartmentalization, the functional roles of these communities remain consistent across the dataset. The network reliably segments into functional groups, such as structural body handlers (Comm 1), boundary/shadow mappers (Comm 2 and 3), and high-frequency texture details (Comm 4). 

Crucially, the Hebbian teacher naturally isolates a large portion of the feature nodes (e.g., 17 nodes). Ablating this entire isolated pool yields a nearly imperceptible difference map, demonstrating empirically that the sparse graph successfully identifies and quarantines redundant or uninformative filters.

\subsection{Micro-Level Explainability via Single-Edge Lesioning}

Beyond macro-level community structures, the S-LooM architecture provides precise, micro-level explainability. To demonstrate that individual graph edges coordinate localized spatial information rather than acting as redundant global passages, we performed single-edge lesioning. 

\begin{figure}[htbp]
    \centering
    \includegraphics[width=1.0\linewidth]{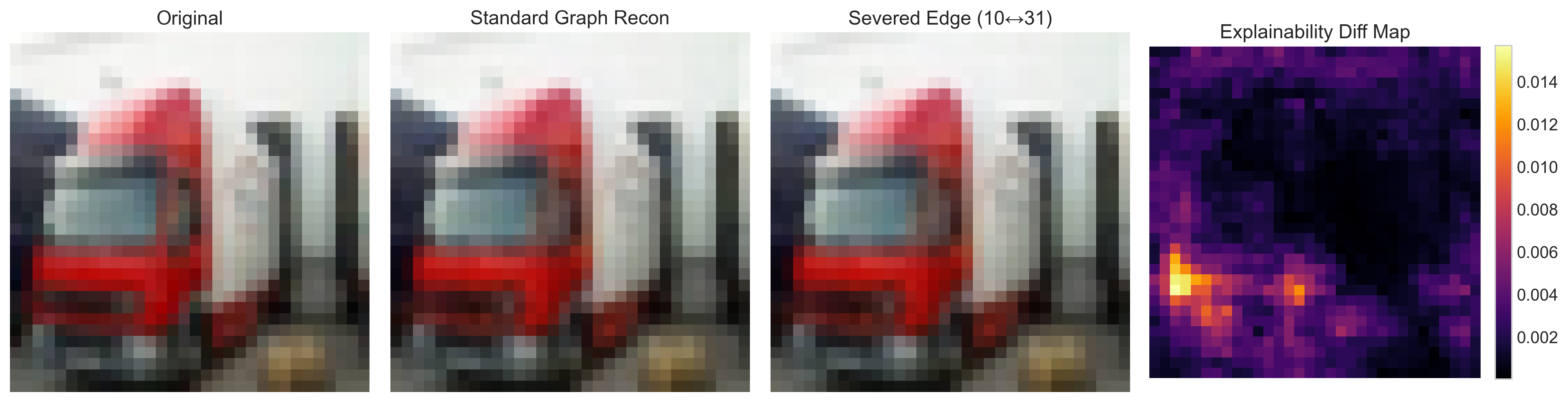}
    \caption{Single-Edge Lesioning. Severing a single graph connection (10 $\leftrightarrow$ 31) results in a highly localized disruption (isolated in the difference map), indicating the network's spatial precision.}
    \label{fig:lesioning}
\end{figure}

As shown in Fig. \ref{fig:lesioning}, severing the single strongest edge in the network (Node 10 $\leftrightarrow$ Node 31) does not cause a generalized blur or a complete channel failure across the reconstructed image. Instead, it strictly degrades a highly localized area, the lower-left headlight and bumper contour of the truck. This demonstrates that S-LooM achieves spatial explainability, allowing us to trace specific visual outputs directly to individual connections within the latent graph.

\subsection{Computational Complexity}

We evaluate the computational overhead of S-LooM by comparing it to a standard convolutional autoencoder. During training, calculating pairwise spatial cross-correlations to construct the Hebbian graph adds a computational cost of $O(K^2 \cdot (H'W')^2)$ per batch, where $K$ is the number of feature maps and $H' \times W'$ is the spatial size of each feature map. To optimize training time, we vectorize this operation using tensor unfolding and parallelized matrix multiplication, effectively trading increased VRAM utilization for rapid GPU execution.

Crucially, because the Hebbian statistics gathering is disabled post-training, the architectural overhead during deployment is minimal. During inference, S-LooM relies solely on the spatial message passing, which mathematically acts as a masked 2D convolution with a complexity of $O(K^2 \cdot s^2 \cdot H' \cdot W')$, where $s$ is the spatial kernel size. Because this operation occurs entirely at the highly compressed latent bottleneck (e.g., $H\times W=8 \times 8$), S-LooM maintains an inference speed almost identical to a standard autoencoder.

\section{Conclusion}

This study introduced S-LooM, a spatially-aware autoencoder architecture, designed to address the limitations of global averaging, scalar message passing and the tendency of latent graphs to entangle under dense connections. By transitioning to localized spatial cross-correlation and learnable spatial kernels, alongside an artifact-free nearest-neighbor decoding scheme, S-LooM mitigates the artificial structural noise (checkerboard patterns) that generated false statistical correlations in prior implementations. Furthermore, a soft-blending training curriculum mitigates representation shock, preventing sudden shifts in latent distributions from disrupting previously learned weights.

Empirical evaluations on CIFAR-10 demonstrate that these architectural refinements yield reduction in baseline reconstruction error. More importantly, the shift from node-based to edge-based sparsity provides stability at high graph densities. Through systematic community ablation and single-edge lesioning, we validated that the resulting sparse latent graph organically compartmentalizes into distinct, explainable functional modules, successfully isolating structural boundaries, high-frequency textures, and redundant connections. Finally, S-LooM demonstrates that with proper spatial grounding, self-organizing Hebbian architectures can transition from theoretical topologies into robust, interpretable representation learning systems. Because this study focuses on isolating the effects of architectural changes compared to the LooM model, evaluations were constrained to the CIFAR-10 dataset. Future work will benchmark S-LooM against a broader array of graph-based models and scale the architecture for higher-resolution, complex datasets to further validate its capabilities.

\end{document}